\documentclass[12pt,journal,compsoc]{article}
\usepackage{graphicx}
\usepackage{amsmath,amssymb} 

\usepackage{geometry}
\usepackage{appendix}
\usepackage{algorithm}
\usepackage{url}

\newtheorem{theorem}{Theorem}[section]
\newtheorem{proposition}[theorem]{Proposition}

\newcommand{\kor}{{k_0}}
\newcommand{\xor}{{x_0}}
\newcommand{\kest}{\widehat{k_0}}
\newcommand{\xest}{\widehat{x_0}}
\newcommand{\xgest}{\widehat{\nabla x_0}}
\newcommand{\xg}{{\nabla x}}
\newcommand{\BE}{\begin{equation}}
\newcommand{\EE}{\end{equation}}
\newcommand{\BEA}{\begin{eqnarray}}
\newcommand{\EEA}{\end{eqnarray}}

\newcommand\restr[2]{{
  \left.\kern-\nulldelimiterspace 
  #1 
  \vphantom{\big|} 
  \right|_{#2} 
  }}

\newcommand\restra[2]{{
#1_{#2}
  }}

\title{Blind Deconvolution with Non-local Sparsity Reweighting} 

\author{Dilip Krishnan$^1$, Joan Bruna$^2$ and Rob Fergus$^{2,3}$ \\
$^1$: CSAIL, Massachussetts Institute of Technology \\
$^2$: Courant Institute, New York University, $^3$: Facebook Inc.}

\begin{document}



\maketitle

\begin{abstract}
Blind deconvolution has made significant progress in the past decade.
Most successful algorithms 
are classified either as Variational or Maximum a-Posteriori ($MAP$). In spite of the superior theoretical
justification of variational techniques, carefully constructed $MAP$ algorithms 
have proven equally effective in practice. In this paper, we show that all successful 
$MAP$ and variational algorithms share a common framework, relying on the following 
key principles: sparsity promotion in the gradient domain, $l_2$ regularization for kernel estimation, 
the use of convex (often quadratic) cost functions and multi-scale estimation. 
We also show that sparsity promotion of latent image 
gradients is an efficient regularizer for blur kernel estimation. Our observations lead to a unified 
understanding of the principles required for successful blind deconvolution. We incorporate these principles
into a novel algorithm that has two new priors: one on the latent image and the other on the blur kernel. 
The resulting algorithm improves significantly upon the state of the art.
\end{abstract}

\section{Introduction}
Starting with the influential work of Fergus et al. \cite{Fergus06}, the state of the art in blind deconvolution
has advanced significantly. For blurred images involving camera translations
or rotations, impressive performance levels have been achieved by a number of algorithms
\cite{Cho09,xu2010two,Goldstein2012,wang2013nonedge,levin2011efficient,whyte11,hirsch11,xuunnatural,wipf2013revisiting}.

The simplest form of the blind deconvolution problem arises
from the following formation model: 

\BE
y = \xor \star \kor + n
\label{eqn:bd}
\EE
where $y$ is the observed blurred and noisy image, $\xor$ the unknown sharp image and
$\kor$ is the unknown blur kernel. 
The noise $n$ is assumed IID Gaussian noise with unknown variance $\sigma^2$. 
Blind deconvolution is the problem of recovering $\xor$ and $\kor$, given
only the observation $y$. The model in \ref{eqn:bd} assumes spatially uniform blur,
and can be extended to non-stationary blurs due to in-plane rotations, as done in Whyte et al. \cite{whyte11}. 
If $\kor$ is known, then the problem reduces to that of non-blind deconvolution 
\cite{Krishnan09a,Levin07}. 

Blind deconvolution is ill-posed since neither the sharp image $\xor$, the
blur kernel $\kor$ or the noise variance are known.
%
To alleviate these issues, prior assumptions on the structure of $\xor$ and $\kor$ 
must be employed. 
A commonly used 
prior on $\xor$ is the heavy-tailed prior (Levin et al. \cite{Levin07}), motivated
from the observation that gradients of natural images follow a hyper-Laplacian distribution. 
Using this prior leads to good results in many applications such as non-blind deconvolution \cite{Krishnan09a}, 
super-resolution \cite{Tappen2003} and transparency separation \cite{Levin07a}. 
If $\nabla x=(\nabla x)_i$, the heavy-tailed distributions used are of the form ${\bf p}(x) = \prod_i p(\nabla x_i)$ with
$p(z) \propto e^{-|z|^\alpha}$
The exponent $\alpha$ is typically in the range of $0.6$ to $0.8$ \cite{Levin07}. 
Priors on the kernel $\kor$ have received lesser attention, but they usually tend to work on
the sparsity of the kernel for motion blurs, such as the $l_1$ norm $\|k\|_1$ (Shan et al. \cite{Shan08}), 
or sparsity of coefficients under a curvlet transform (Cai et al. \cite{cai2009blind}).

Unfortunately, using the above priors in a naive alternating minimization 
(AM) framework leads to the trivial solution 
$\xest = y, \kest = \delta$, where $\delta$ is the Dirac. 
In \cite{Levin09}, Levin et al.
analyze the reasons behind this phenomenon, when the heavy-tailed prior is used. 
The fundamental reason is quite simple: the probability of a sharp image $x$ is lower
under the commonly used heavy-tailed prior, with exponent in the range of 
$0.6$-$0.8$. In their paper, Levin et al. also identified a workaround. The same authors, in 
a follow-up work \cite{levin2011efficient} present a simplified version of the algorithm of  
Fergus et al. \cite{Fergus06}. The paper \cite{Fergus06} itself was based on the work 
of Miskin and MacKay \cite{miskin2000ensemble}.
We call this family of related algorithms the variational model.


A different family of algorithms such as those of \cite{xu2010two,Cho09} are 
categorized as Maximum a-Posteriori ($MAP$). The chief distinction between the variational and $MAP$ algorithms is the 
use of probability distributions in the former, as opposed to point estimates in the latter.
The kernel estimate $\kest$ is thus obtained
by marginalizing the posterior 
distribution over all possible images $x$. This Bayesian approach is usually seen as a 
strong advantage for the variational methods since the uncertainty of an estimate is taken into
account. Indeed, they perform well empirically. 
However, in practice, the marginalization 
is intractable and a series of approximations are performed to realize a practical algorithm. 
$MAP$ formulations, on the other hand, use alternating minimization (AM) updates on $\xest$ 
and $\kest$, resulting in non-convex optimizations. 
In spite of this seemingly inferior formulation, in practice the best $MAP$ formulation techniques
have proven as effective as variational methods. 
The key to their performance is the use of
additional steps to supplement the AM iterations. 

We make the following contributions in this paper: 
we first show that the use of approximations in the variational method 
and non-naive approaches in $MAP$ methods lead to essentially the same framework. 
We show theoretically and experimentally that sparsity inducing regularizations are the key ingredient, irrespective
of whether they provide good image gradient priors or not.
This helps explain why the top-performing methods all achieve similar performance.
We develop a new algorithm based on our insights that achieves significantly better than state of the art results 
on the standard benchmark of \cite{Levin09}. Our algorithm has two novel priors. The first is a patch-based
sparsity promotion prior on the latent image which estimates salient geometric features that are 
crucial for good kernel estimation. The second is a frequency-domain based kernel prior that 
performs adaptive regularization of the latent kernel to alleviate the aperture problem.

Our work has shared ground with that of Wipf and Zhang \cite{wipf2013revisiting}, who also seek to explain
the reasons behind the success of the variational approach. 
We show that most successful algorithms (not just variational) follow similar principles.  
Our resulting recipes are conceptually simpler than that suggested by \cite{wipf2013revisiting}, and
we also provide directions for future improvements.

The variational and $MAP$ paradigms do not cover all deconvolution algorithms.
For example, the spectral analysis based algorithm of Goldstein and Fattal \cite{Goldstein2012} and the Radon transform 
based method of Lin et al. \cite{chohandling} are two examples where our current analysis does not hold. 
Nevertheless, we note that at present, these alternative methods do not perform at state of the art levels compared
to the $MAP$ and variational algorithms.

{\it Notations:} We denote by $\mathcal{F}(x)$ or $x_f$ the Fourier transform of $x$.
$\nabla x = (\partial_{h} x, \partial_{v} x)$ denotes the gradient of a two-dimensional
signal.

\section{Variational and $MAP$ Approaches}
\label{sec:review}
In this section, we consider in detail the variational algorithm of Levin et al. \cite{levin2011efficient} and Wipf and Zhang\cite{wipf2013revisiting}, 
and the $MAP$ algorithms of Xu et al. \cite{xuunnatural}, Xu and Jia \cite{xu2010two} and Cho and Lee \cite{Cho09}. 
These algorithms are all considered state of the art, and perform very well on the 
benchmark dataset of Levin et al. \cite{Levin09}. 

All of the above algorithms work in the gradient domain for kernel estimation. 
Since convolution commutes with derivatives, this does not change the form of the cost function \ref{eqn:bd}. 
The gradient space is used to determine a kernel $\kest$, and the final sharp 
image $\xest$ is typically recovered with a non-blind deconvolution algorithm such as Krishnan and Fergus \cite{Krishnan09a}. 

\subsection{Naive $MAP$}
\begin{figure}[t]
\begin{center}
\includegraphics[width=3in]{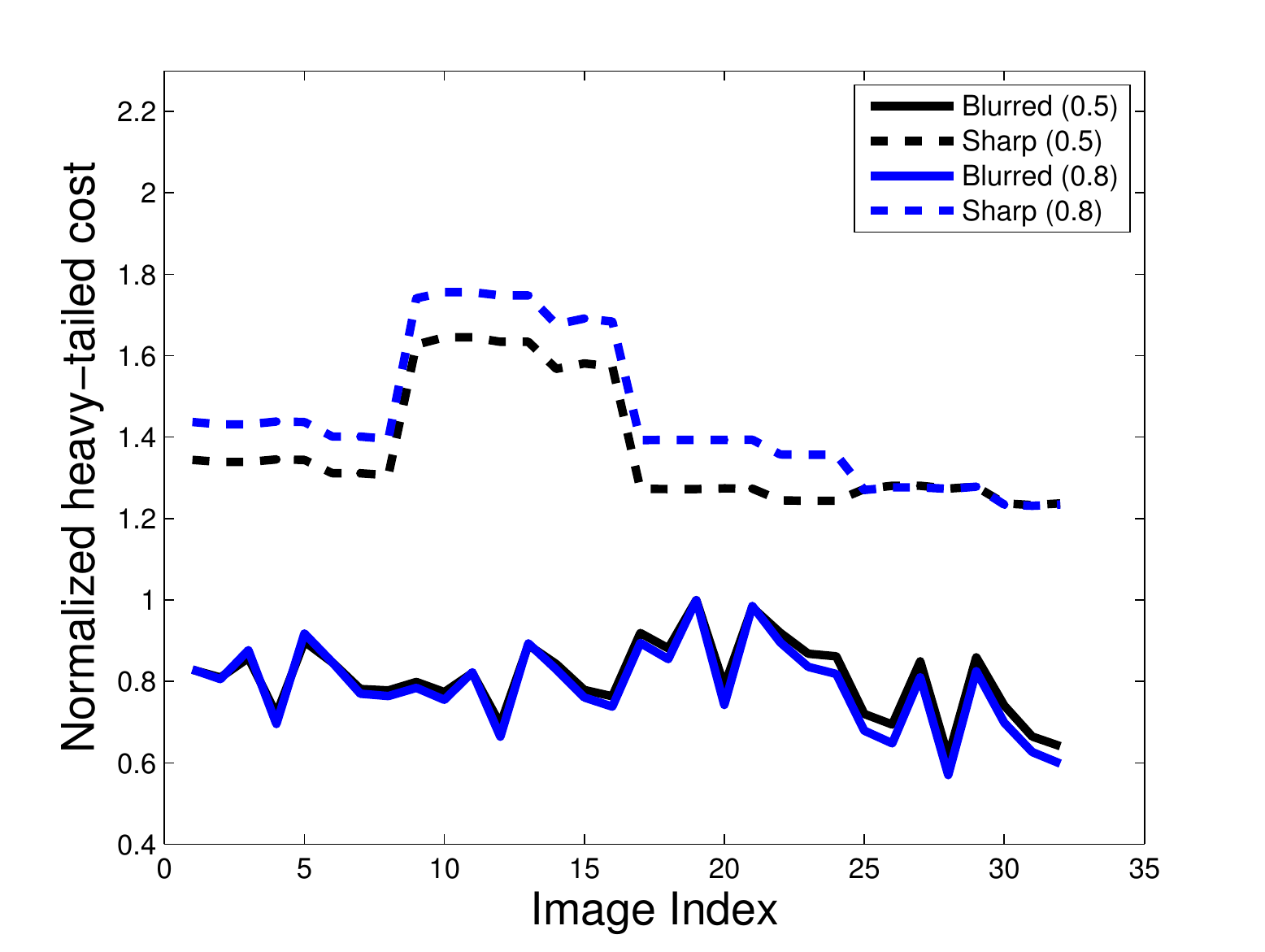}
\end{center}
\vspace{-5mm}
\caption{Comparison of costs of blurred and sharp images under heavy-tailed prior: 
$32$ images from the dataset of Levin et al. \cite{Levin09} for $\alpha = 0.5$ and $\alpha = 0.8$. Gradients
of blurred images have lower cost. }
\label{fig:lp_cost}
\end{figure}

The naive $MAP$ algorithm that is prone to poor solutions solves the following
cost function:
\BE
(\xgest, \kest)  = \arg \min_{\xg,k} \lambda \| \nabla y - \xg \star k \|^2 + \sum_i |\xg_i|^\alpha
\label{eqn:naive_map}
\EE
Alternating minimization is usually employed: given a current 
estimate ${k}_n$, a new update ${\xg}_{n+1}$
is computed, and vice-versa. The regularizer on $\xg$ 
is a heavy tailed prior Levin et al. \cite{Levin09} with $\alpha < 1$.  
It has been shown in \cite{Levin09} that this cost function 
leads to the trivial solution $\xest = y, \kest = \delta$. 
This is because the trivial solution achieves the 
lowest cost for both the likelihood term $\|\xg \star k - \nabla y\|^2$
and the regularizing term $\sum_i |\xg |_i^\alpha$. 
\ref{fig:lp_cost} shows this phenomenon for the 
$32$ blurred images from the dataset of \cite{Levin09} for values of $\alpha = 0.5$ and $\alpha = 0.8$. 
Heavy-tailed priors
give a lower cost to the blurred image because the blurring operation reduces the overall gradient
variance, which reduces $\sum_i |\xg_i|^\alpha$. 
On the other hand, because zero gradients near strong edges
become non-zero due to blur, an opposite effect is that $\sum_i |\xg_i|^\alpha$ is increased by blurring.
For $\alpha = 0.5$ or larger, the former effect dominates and this causes the measure to prefer
the blurred image. It is shown in Wipf and Zhang \cite{wipf2013revisiting}, that for very small $\alpha$ values, the
situation may be reversed. However, the resulting cost functions are numerically unstable and 
difficult to handle.

\subsection{Successful $MAP$ Methods}
In Cho and Lee \cite{Cho09}, alternating $x$ and $k$ updates are performed using the following equations:
\BEA
{x}_{n+1} =\arg \min_x \sum_{j} \| \partial_{j} x \star {k}_n - \partial_{j} y\|^2 + \alpha \| \partial_{j} x\|^2 \nonumber \\
{k}_{n+1} = \arg \min_k \sum_{j} \| \partial_{j} {x}_{n+1} \star k - \partial_{j} y\|^2 + \beta \| k\|^2~,
\label{eqn:cho09}
\EEA
where $j$ indexes a set of partial derivative filters; in their implementation \cite{Cho09} use $6$ filters 
\footnote{The filters are first-order and second-order derivative filters in horizontal, vertical and diagonal
directions.}. 
Clearly, due to the phenomenon seen in  \ref{fig:lp_cost}, this simple formulation has little hope of succeeding since
the quadratic regularization forces ${x}_n$ towards the blurred image $y$. Therefore, 
Cho and Lee introduce an additional step to promote sparsity in  $\{\partial_{\gamma} x\}$. 
This additional step is a shock filter Osher and Rudin \cite{osher1990feature}, 
which suppresses gradients of small magnitude and 
boosts large magnitude gradients. This shock filtering step is performed
after the $x$ update step, and prior to the $k$ estimation, 
thereby preventing a drift towards the trivial solution.

Xu and Jia \cite{xu2010two} also use a shock filter, and additionally an importance map, which is designed to 
down weight the importance of low magnitude gradients as well as isolated spikes. 
The $k$ update step is identical to that of Cho and Lee \cite{Cho09}, and is given in \ref{eqn:cho09}, 
also using an $l_2$ (quadratic) norm on $k$.

The very recent work of Xu et al. \cite{xuunnatural} employs an $\ell_0$-like prior on $\xg$. 
The cost
functions that they solve to update $x$ and $k$ are given by:
\BEA
{x}_{n+1} &=& \arg \min_x \| y - x \star {k}_n \|^2 + \lambda \Phi (\nabla x) \nonumber \\
{k}_{n+1} &=& \arg \min_k \| y - {x}_{n+1} \star k \|^2 +  \gamma \| k\|^2 ~,
\label{eqn:xu2013}
\EEA
where $\Phi$ is a function that approximates  $\| \nabla x\|_0$. 
The $x$ update step involves a series of quadratic relaxations that progressively approximate the $\ell_0$ 
function more closely, thereby imposing sparsity on the gradients $\xg$.
The above papers, \cite{Cho09,xu2010two,xuunnatural} and other $MAP$ methods, 
periodically enforce non-negativity and sum-to-1 constraints of the entries of $k$. Generally, this 
is done after a $k$-update step. 

\subsection{Variational Methods}
The variational method was introduced to blind deconvolution by Miskin and Mackay \cite{miskin2000ensemble}, 
who considered the blind deconvolution and separation of cartoon images. They 
imposed a highly sparsity promoting prior on $x$, given by a mixture of Laplacians.

Fergus et al. \cite{Fergus06} extended the algorithm of
Miskin and Mackay to natural images. The contribution of \cite{Fergus06} was to realize that the gradient fields
of natural images are generally highly sparse, and thereby working in gradient space
allows the extension of the original algorithm of Miskin and Mackay.

A conceptually simpler version of the algorithm of \cite{Fergus06} was given by Levin et al. \cite{levin2011efficient}. 
While \cite{Fergus06} is variational in both $x$ and $k$,
\cite{levin2011efficient} is variational only in $x$. 
Under a probabilistic interpretation of blind deconvolution, the 
estimation of $k$ is given by:
\BE
\hat{k} = \arg \max_k p(k | y) = \arg \max_k \int p(y|k,x)p(x) dx
\label{eqn:map_k}
\EE

However, \ref{eqn:map_k} is computationally intractable, 
and variational approximations are introduced in \cite{Fergus06,levin2011efficient}
to realize a practical algorithm. One can show \cite{wipf2013revisiting,levin2011efficient} that 
the final form of the resulting algorithm has the form
{\small 
\BEA
{\xg}_{n+1} &=& \arg \min_x \frac{1}{\eta_n^2} \| \nabla y - x \star {k}_n \|^2 + \sum_i ( w_{i,n} x)^2 \label{eqn:eqxvar}
 \\
{k}_{n+1} &=& \arg \min_k \| \nabla y  - {\xg}_{n+1} \star k \|^2  + \lambda_n \| k \|^2 ~,
\nonumber
\EEA
}
where $\eta_n$ refers to a noise level parameter and
the weights $w_{i,n}$ evolve dynamically to penalize current estimates $\xg_{i,n}$
of low gradient amplitudes and to ``protect" large gradients. 
The resulting iterative minimization therefore favors a sparse $\xg_{n+1}$. 
Note that in \ref{eqn:eqxvar}, we are estimating the latent image gradients.
On the other hand, the $k$ step consists of a ridge regression, 
where the parameter $\lambda_n=\mbox{Tr}(\Sigma_{n}^{-1})$ 
and $\Sigma_{n}$ is a diagonal covariance of ${\xg}_n$ 
estimated from the previous $x$-step \cite{levin2011efficient}. 
As a result, the regularization 
strength is a measure of the overall variance in the estimate of $\xg_{n+1}$.

\section{The Common Components}
This section explains why sparsity promoting regularizations
play a central role for blind deconvolution. We argue 
that the main reason is not related to the prior distribution of image gradients.

\subsection{Sparsity Promotion}
\label{sect_xdisc}
The total variation has been extensively used 
as an efficient regularizer for several inverse problems 
\cite{chambolle2004algorithm,Rudin92}, including denoising and 
non-blind deconvolution. 
It corresponds to the $\ell_1$ norm computed on image
gradients $\nabla x$, which is known Rudin et al. \cite{Rudin92} to promote 
 solutions whose gradients are sparse.

This suggests that a similar sparsity-promoting prior
will also be useful for the blind-deconvolution inverse 
problem. For that purpose, several authors \cite{Shan08,chan1998total} 
suggested using $\| \nabla x \|_p$ with $p \leq 1$ as a prior.
Similarly, all variational approaches are based on 
sparsity promoting priors Wipf and Zhang \cite{wipf2013revisiting}. 
Since the derivative is a linear, translation invariant 
operator, we have $\nabla y = (\nabla \xor) \star \kor + \nabla n$. 
 This results in a cost function of the form
$$\| \nabla y - x \star k \|^2 + \Phi(x)~,$$
where $\Phi$ is a sparsity-promoting function. 
Since natural images typically have a spectrum 
decaying as $\sim \omega^{-2}$ and $\mathcal{F}(\partial x)(\omega)=i \omega \mathcal{F}(x)(\omega)$, 
it results that the likelihood term
expressed in the gradient domain is simply a reweighted 
$\ell_2$ norm with equalized frequencies. 

However, the blind deconvolution inverse problem 
requires not only the estimation of $x_0$ but also  
estimating the kernel $k_0$. We argue that enforcing sparsity of
$\nabla x$ is a regularizer for $\kest$ which is highly efficient,
even when input images do not have sparse gradients.

We shall consider a ridge regression ($l_2$ norm) on the kernel.
Let us concentrate on the case of spatially uniform blur of
\ref{eqn:bd}, and let us suppose the kernel $k_0$ has 
compact support of size $S$.
 The following proposition, 
 proved in Appendix \ref{proofsect}, 
 shows that if one is able 
 to find an approximation of $\nabla \xor$ which has 
 small error in \emph{some} neighborhood $\Omega$ 
 of the image domain, then setting to zero $\nabla x$ 
 outside $\Omega$ yields a good approximation 
 of $\kor$. We denote $dist(i, \Omega) = \inf \{ |i - j|\,,\, j \in \Omega\}$.
 \begin{proposition}
 \label{sparseprop}
Let $y=\xor \star \kor + n$, with $\sum_i \kor_i =1$.
For a given $x$ and a given neighborhood
$\Omega$, let
\begin{eqnarray}
\epsilon^2 &=& \| x - \xor \|_{\Omega,S}^2 := \sum_{dist(i,\Omega) \leq S} |x_i - \xor_i|^2 ~, \nonumber \\
\gamma^2 &=&  \| \xor \|_{\Omega,S}^2 ~,
\end{eqnarray}
and let us assume that the matrix 
$A$ whose columns are 
$$(A)_j = \{ \xor_{j-i} \, ; |i| \leq S \}_{j \in \Omega}$$
satisfies $\lambda^2_{\min}(A)=\inf_{\sum_i y_i =0\,,\,\|y\|=1} A(y)= \delta >0$.
Then, by setting 
\begin{equation}
\tilde{x}_i = \left \{ 
\begin{array}{ll}
x_i & \mbox{if } i \in \Omega~, \\
0 & \mbox{otherwise}~,
\end{array} \right.
\end{equation}
the solution of 
\begin{equation}
\kest = \arg \min_{k \, s.t. \sum_i k_i=1} \| y - \tilde{x} \star k \|^2 + \lambda \| k \|^2
\end{equation}
satisfies 
\begin{equation}
\label{egg3}
\| \kest - \kor \| \leq C \| \kor \| + c~,
\end{equation}
where $C=O(\max(\epsilon \gamma \delta^{-1},\lambda))$ and $c=O(\|n\|_{\Omega}\gamma \delta^{-1})$.
 \end{proposition}

This proposition shows that in order to recover a good estimation of the kernel, it is sufficient to 
obtain a good estimation of the input gradients on a certain neighborhood $\Omega$. 
Sharp geometric structures and isolated singularities are natural candidates to become part of $\Omega$, 
since they can be estimated from $y$ by thresholding the gradients. This partly explains the numerical
success of shock filtering based methods such as those in Cho and Lee \cite{Cho09} and Xu and Jia \cite{xu2010two}. 

Promoting sparsity of the image gradients thus appears to be an efficient mechanism to identify 
the support of isolated geometric features, rather than a prior for the distribution of image gradients.
In particular, Proposition \ref{sparseprop} shows that images having textured or oscillatory regions 
do not necessarily increase the approximation error, as long as they also contain geometric features.
Proposition \ref{sparseprop} gives a bound on the estimation error of $\kor$ given a local approximation of $\xor$. 
The error is mainly controlled by $\epsilon$, the approximation error of $\xor$ on the active set $\Omega$, and
$\delta$, which depends upon the amount of diversity captured in the active set. The so-called
\emph{aperture problem} corresponds to the scenario $\delta=0$, in which $\kor$ can be recovered only on the subspace
spanned by the available input data.

Finally, let us highlight the connection between 
this result and the recent work of Ahmed et al. \cite{ahmed2012blind}: 
the authors show that under 
certain identifiability conditions, one can recover 
$\xor$ and $\kor$ by solving a convex program on 
the outer product space. In this sense, 
the sparsity 
enhancement of $x$ helps identify a subspace $\Omega$
such that the restrictions $\restr{y}{\Omega}$, $\restr{x}{\Omega}$ satisfy
better identifiability conditions.

\subsection{$\ell_2$ norm on $k$}
\label{sect_kdisc}
The inverse problem of \ref{eqn:bd} requires regularisation 
not only for the unknown image but also for the unknown kernel.
It is seen from \ref{sec:review} that all the top-performing methods 
use an $\ell_2$ ridge regression on the kernel $k$, which 
regularises the pseudo inverse associated to  
$$\min_k \| \nabla y - \nabla \hat{x} \star k \|^2~.$$ 
An $\ell_2$ norm gives lower cost to a diffuse kernel, 
which helps to push away from the trivial solution $k = \delta$. 
Moreover, the previous section showed that
the necessary sparse regularisation of the $x$-step may cause
the regression to be ill-conditioned due to the aperture
problem. 

Since the ridge regression only contains Euclidean norms, 
one can express it in the Fourier domain
$$\min_k \| y_f - x_f \cdot \mathcal{F}(k) \|^2 + \lambda \| \mathcal{F}(k) \|^2~,$$ 
where $y_f$ and $x_f$ are respectively the Fourier transforms of $\nabla y$ and $\widehat{\xg}$ 
computed at the resolution of the kernel. It results 
in the well-known Wiener filters, in which  
frequencies with low energy in the current estimate $\widehat{\xg}$ 
are attenuated by the ridge regression. 
This may create kernels with irregular spectra, which 
translates into slow spatial decay, thus producing diffused 
results. In order to compensate for this effect, 
some authors such as Levin et al. \cite{levin2011efficient} introduced
a sparsity-promoting term in the estimation of $k$ as well. 
Since we assume positive kernels with constant DC gain 
(set to $1$ for simplicity), $\| k \|_1 =1$ by construction, 
thus requiring a regulariser of the form $\| k \|_p$ with $p<1$ in 
practice.


\subsection{Convex Sub-problems}
A notable aspect of the successful algorithms is the use of quadratic cost functions for both
the $x$ and $k$ sub-problems (even though the joint problem is non-convex). 
Quadratic cost functions are especially simple to optimize when 
convolutions are involved: fast FFT or Conjugate Gradient methods may be used. 
For non-quadratic convex cost functions, iteratively reweighed least
squares Daubechies et al. \cite{daubechies09} may be used.

When using a convex sparsity-promoting regularizer for $\xg$, 
one may compromise the sparsity promotion ability.
However, this must be balanced against the fact that for a non-convex regularizer, it can be hard to achieve 
a sparse enough solution, as 
seen in the results of Krishnan et al. \cite{krishnan2011blind}, which 
uses a non convex regulariser. 

The tradeoff between sparsity-promotion and the solvability of a regularizer is therefore an important 
design criterion. The re-weighted methods of Levin et al. \cite{levin2011efficient} and Xu et al. \cite{xuunnatural} seem to strike
a good balance by solving convex (quadratic) cost functions. 
In our experiments with the publicly released code of \cite{levin2011efficient}, we found that solving each sub-problem
to a high level of accuracy was crucial to the performance of the method. For example, reducing the 
number of conjugate gradients iterations in the $\xg$ update of \ref{eqn:eqxvar}. 
caused the performance to be much poorer. 
This is due to the lack of sufficient level of sparsity in the resulting $\xg$. 

\subsection{Multi-scale Framework}
Due to the non-convex nature of the blind deconvolution problem, it is easy to get stuck at a local
minimum. A standard mechanism to overcome this is to use a coarse-to-fine framework for estimating the kernel.
This coarse-to-fine scheme is used by all successful algorithms. At each scale in the pyramid, the upsampled kernel
from the coarser level, and the downsampled blurred image from the finest level are used as an initialization. 
At the coarsest level, a simple initialization away from the $\delta$ kernel is used, such as a 2-pixel horizontal or
vertical blur. 

\section{Our New Algorithm}
\label{sec:ours}
We combine the principles described above into a new algorithm that performs above the state of the art on the 
benchmark dataset of Levin et al. \cite{Levin09}. In addition to the high performance, an advantage
of our method is that it has only two user-defined parameters that determine the regularization
levels on the estimation of $k$. This is in contrast with methods such as \cite{xuunnatural,Cho09} which have 
a few parameters whose settings can be hard to estimate. 

We work in derivative space, using horizontal and vertical derivative filters. 
As argued in section \ref{sect_xdisc}, 
our $x$ update step is given by a reweighted least squares formulation which 
promotes solutions with isolated geometric structures,
whereas the $k$ update solves a least squares regression using $\ell_2$ and $\ell_p$ 
regularisatio discussed in \ref{sect_kdisc}. However, unlike in \ref{sect_kdisc}, we use a novel reweighted $\ell_2$ prior on $k$ 
(discussed below):
{\small
\BEA
{\xg}_{n+1} &=& \arg \min_x \| \nabla y - x \star {k}_n \|^2 + \sum_i w_{i,n} x_i^2~,
\label{eqn:our_x} \\
{k}_{n+1} &=& \arg \min_k \| \nabla y - {\xg}_{n+1} \star k \|^2  +  \| A k \|^2_2 + \lambda \| k \|_{0.5}~.
\nonumber 
\EEA 
}
The weights $w_{i,n}$ at each iteration are based on the current estimate ${\xg}_n$. 
They are designed to select the regions of $\xg_n$ with salient geometrical features while
attenuating the rest. 
Let $p_{i,n}$ be the patch of size $R$ centered at pixel $i$ of ${\xg}_n$.
We consider 
\BE
w_{i,n} = \frac{\eta}{\eta+ |{\xg}_{i,n}| \cdot \| p_{i,n} \|_2 }~. 
\label{eqn:our_wts}
\EE
%
The values of $w_{i,n}$ range between $0$ and $1$, and  
they are inversely proportional to $|{\xg}_{i,n}|$. 
Small gradients will have a larger regularization weight (close to $1$),
and as a result these small gradients will tend to be shrunk towards $0$ in \ref{eqn:our_x}.
However, point-wise reweighting does not have the capacity to separate geometrically 
salient structures, such as edges or isolated singularities, from textured regions.
Proposition \ref{sparseprop} showed that isolated gradients, corresponding
to those salient geometric features, provide better 
identifiability than regions with dense large gradients. 
In order to perform this geometric detection, it is thus necessary to consider 
non point-wise weights. \ref{eqn:our_wts} considers the local $\ell_2$
norm $\| p_{i,n} \|_2$ over a neighbourhood at each given location. 
Isolated features have large local energy relative to 
 non-sparse, textured regions. Therefore, $w_{i,n}$ will
tend to attenuate those textured regions in favour of salient geometry.
In our experiments, we set patch size $R=5$ and $\eta=\| \nabla y - \nabla x_n \star k_n \|^2$ 
to progressively anneal the offset in \ref{eqn:our_wts}.


Our $k$-update step uses a sparsity promoting $\ell_p$ norm $\lambda \| k \|_{0.5}$ with 
$\lambda = 6 \cdot 10^{-3}$. We also introduce a novel reweighted ridge regression prior on the kernel.
The standard unweighted ridge regression term $\| k \|_2^2$ acts uniformly on 
all frequencies of the kernel $k$, since $\|k\|^2 = \| \mathcal{F} k\|^2 = \sum_{\omega} |k_f (\omega)|^2$, 
where $k_f \equiv \mathcal{F}k$.
We change this to a \emph{frequency dependent} weighting $\sum_{\omega} \alpha_{\omega} |k_f (\omega)|^2$.
The positive weights $\alpha_{\omega}$ are chosen to counteract the effect of aperture in the 
blurring process.  

When a certain frequency of the observation $y_f(\omega)$ has very little energy, there is a fundamental
ambiguity: is $|y_f(\omega)|$ small because $|x_f(\omega)|$ was near-zero (aperture)
or was there a near-zero in the frequency of the kernel $k_f(\omega)$ that attenuated the energy
in $\hat{x}(\omega)$? Hence at such ambiguous frequencies, we increase regularization strength. On other frequencies
with significant energy, we reduce the regularization strength. We therefore choose the weights $\alpha_{\omega}$ to be 
inversely proportional to the observation energy, as follows: 

\BE
\alpha_{\omega} = \frac{\lambda_{ap}}{1 + |y_f(\omega)|}
\label{eqn:kernel_aperture}
\EE

where $\lambda_{ap}$ is set to $200$ for all experiments reported in this paper. Note that we are
using the Fourier transform of the observed image $y$ and not the Fourier transform of $\nabla y$. A constant of $1$ was 
empirically added to the denominator to give a certain minimal level of regularization for robustness to 
noise and to move away from the trivial $\delta$ kernel. The matrix $A$ given in
\ref{eqn:our_x} above is therefore simply a product of a diagonal matrix with the appropriate Fourier matrix. The diagonal
entries of the diagonal matrix are given by $\sqrt{\alpha_{\omega}}$. 

We solve the $x$ update step in (\ref{eqn:our_x}) by performing $30$ iterations of Conjugate Gradient with a fixed
value of weights $w_{i,n}$, which
achieves high accuracy owing to its quadratic formulation.
The kernel update in (\ref{eqn:our_x}) is solved using IRLS.  
 After every $k$ update, we set negative elements
of $k$ to $0$, and normalize the sum of the elements to $1$. We embed the entire framework 
in a multi-scale framework and perform
$20$ alternating iterations of $x$ and $k$ at each level. The weights $w_{i,n}$ are updated
after every alternating iteration. The weights $\alpha_{\omega}$ and the matrix $A$ are computed
once at the beginning and do not change during the iterations. 

\section{Experimental Results}
In this section, we compare our algorithm to that of Cho and Lee \cite{Cho09}, Levin et al. 
\cite{levin2011efficient}, and Xu et al. \cite{xuunnatural}. 
Our algorithm parameters are fixed to the values given in \ref{sec:ours}.

We start with the test dataset of \cite{Levin09}. This consists of $4$ images blurred with
$8$ motion blur kernels, giving rise to $32$ blurred image-kernel pairs. The standard method
of comparison is to compute the ratio of the mean square error of the recovered image with the mean square
error of the blurred image deconvolved with the ground-truth kernel, which is known. For all comparisons
in this section, we use the sparsity based non-blind deconvolution method of Levin et al. \cite{levin2011efficient} to perform
the final non-blind deconvolution step. We use the executable downloaded from the website of the authors of 
\cite{xuunnatural} and used existing results for \cite{Cho09} (provided with the code of \cite{levin2011efficient}). 
We used the same non-blind deconvolution technique provided with the
code of \cite{Levin09} with the same parameter settings. 

\begin{figure*}
\includegraphics[width=2.3in]{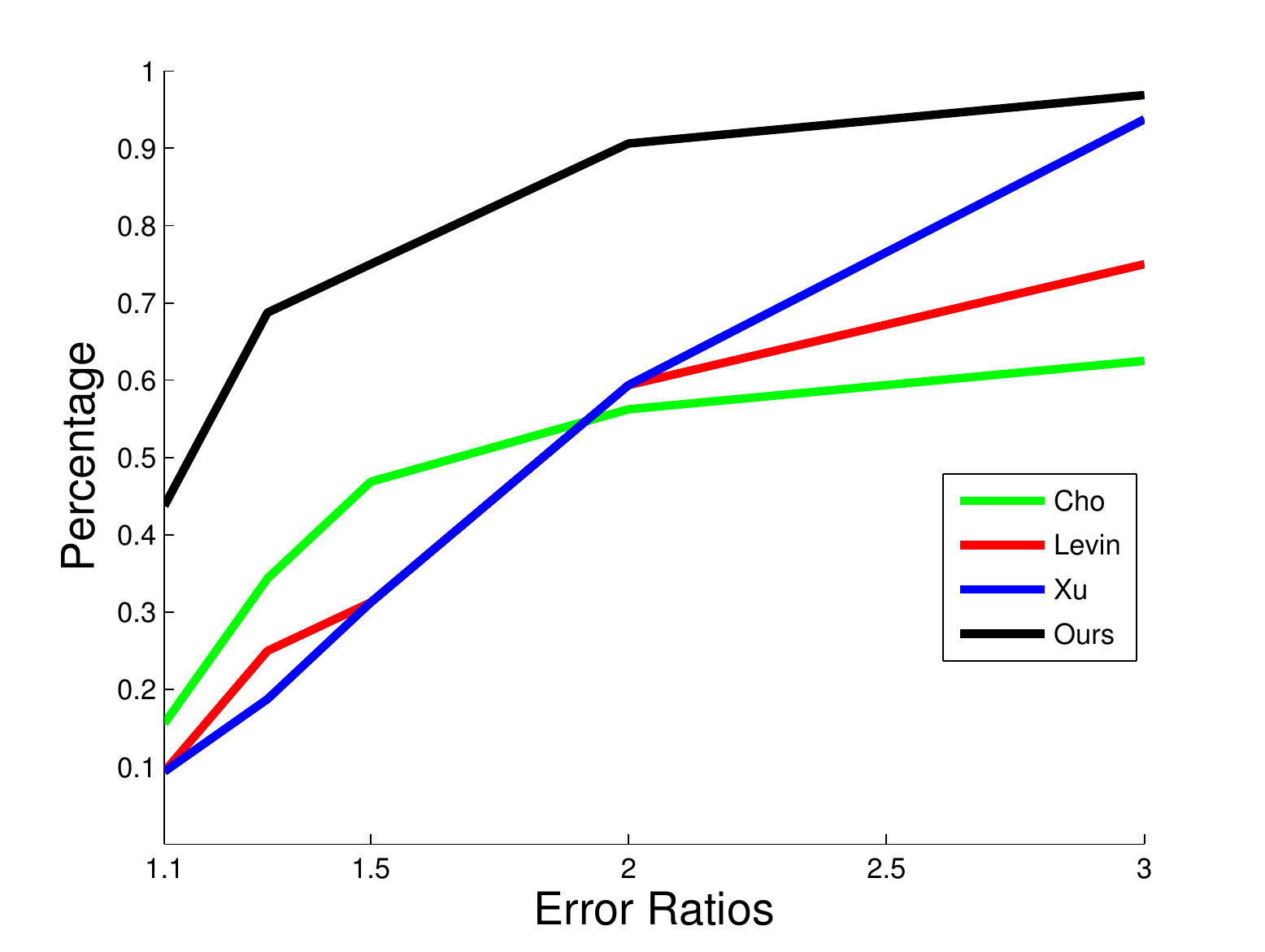}
\hspace{3mm}
\includegraphics[width=2.3in]{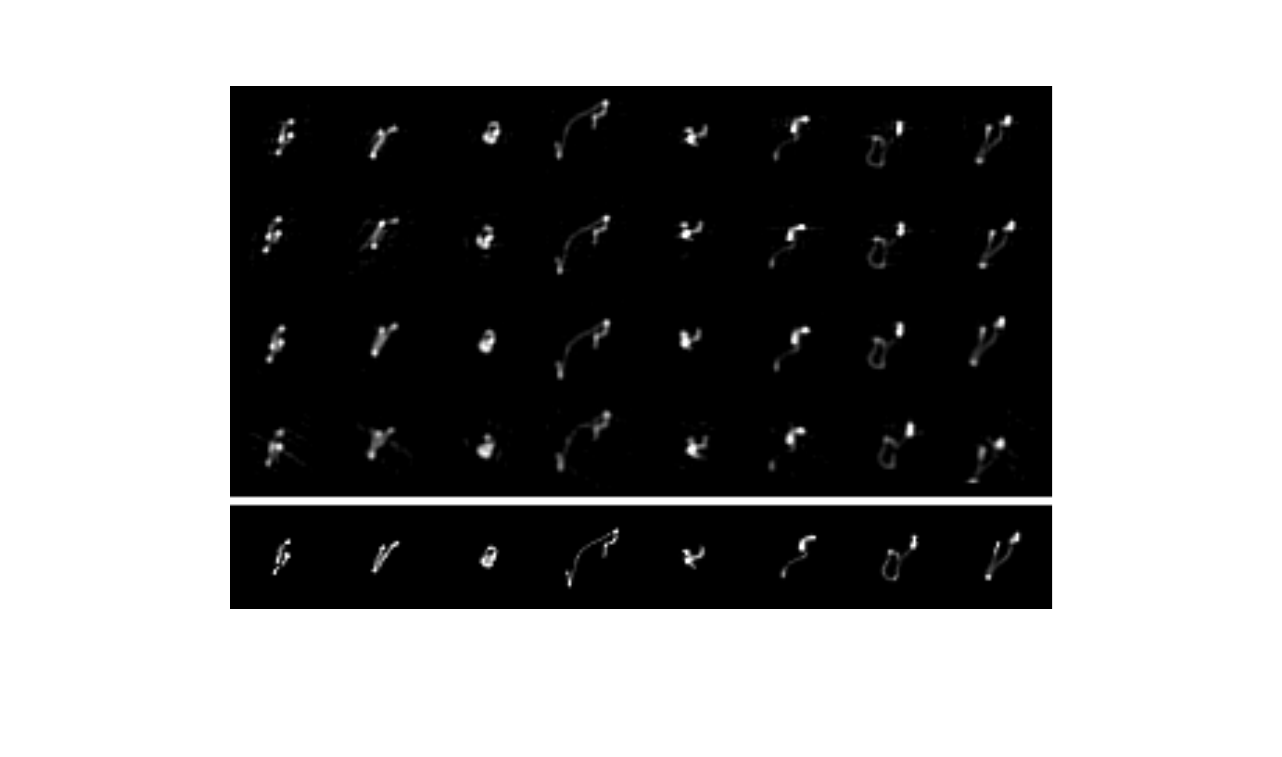}
\vspace{-5mm}
\caption{Left: Performance on dataset of Levin et al. \cite{Levin09}. We compare the following methods: the variational algorithm of 
Levin et al. \cite{levin2011efficient}; the $MAP$ algorithms of Cho and Lee \cite{Cho09} and Xu et al. \cite{xuunnatural}; 
and our new algorithm. Our algorithm is the top-performing. Right: our recovered kernels are shown: the top $4$ 
rows correspond to the $4$ images and the $8$ columns correspond to the kernels we recover for each image. The 
last row shows the $8$ ground truth kernels.}
\label{fig:anat_database}
\end{figure*}

Error ratios less than $3$ are considered visually good. \ref{fig:anat_database} shows the cumulative error ratios
and our recovered kernels for the different images. It is seen that
our algorithm outperforms the other methods, with 
$90\%$ of the images achieving an error ratio less than $2$.
 However, all the algorithms perform quite well.
This is to be expected since each of these methods does promote 
sparsity of the gradients. 
The kernels we recover, are very close to the ground-truth kernels shown in the 
last row. Using patches of size $R = 5$ in 
\ref{eqn:our_wts} performed better than using $R = 1$ i.e. point wise estimation. This is 
not surprising as isolated gradients can be better detected with larger patch sizes.

The effect of the new kernel prior (\ref{eqn:our_x}) is especially obvious in the fourth
image from the dataset of \cite{Levin09}. This image consists of most gradients oriented
in a particular direction and very few in other directions. This leads to 
poor estimation for frequencies that are orthogonal to the dominant frequencies. 
The use of an isotropically weighted kernel $\ell_2$ prior leads to either an excessively
diffuse kernel in low energy directions or an excessively sparse kernel in all directions, depending
on the regularization strength. In fact, this artifact is visually visible in the results 
of both Xu et al. \cite{xuunnatural} and Levin et al. \cite{levin2011efficient}, since they use an isotropically weighted
$\ell_2$ prior. Numerically as well, the errors are higher for this image. 
In our results in \ref{fig:anat_database}, the kernels on the fourth row are visually and numerically nearly
as accurately recovered, as for the other rows.

%

\begin{figure*}
\begin{center}
\includegraphics[width=4in]{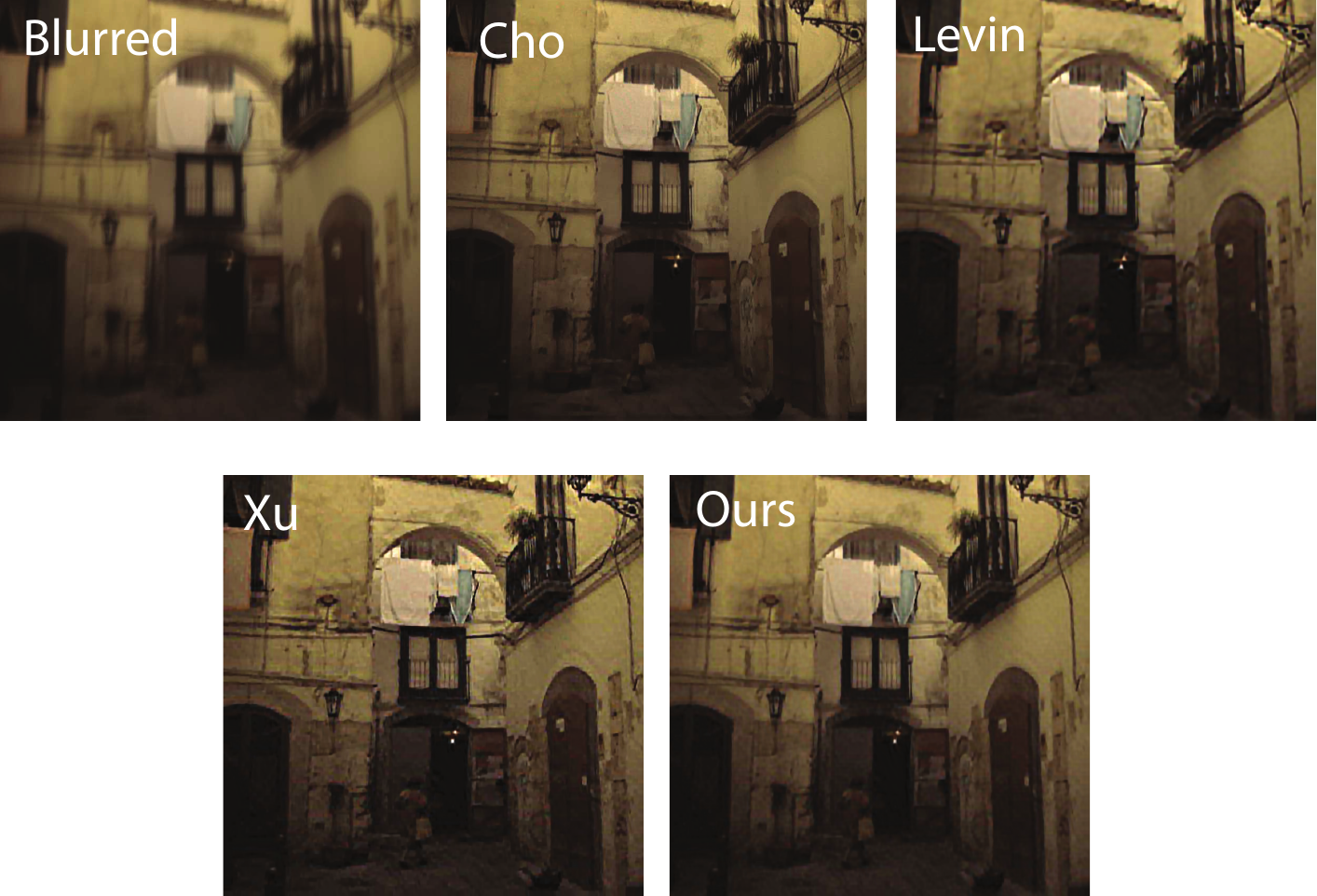}
\end{center}
\vspace{-4mm}
\caption{A real-world example from Xu et al. \cite{xuunnatural}. Our method performs as well as
that of \cite{xuunnatural} and Cho and Lee \cite{Cho09}. Levin et al. \cite{levin2011efficient} exhibits some ringing artifacts.}
\label{fig:real_xu}
\end{figure*}

\vspace{-3mm}
Next, we compare with some real-world examples. In \ref{fig:real_xu}, we compare methods on an 
example from Xu et al. \cite{xuunnatural} (distributed as part of their software package). We show here the output of the
executable of \cite{xuunnatural}, which appears somewhat inferior to the result in their paper (nevertheless still
being quite good). 

\begin{figure*}
\begin{center}
\includegraphics[width=4in]{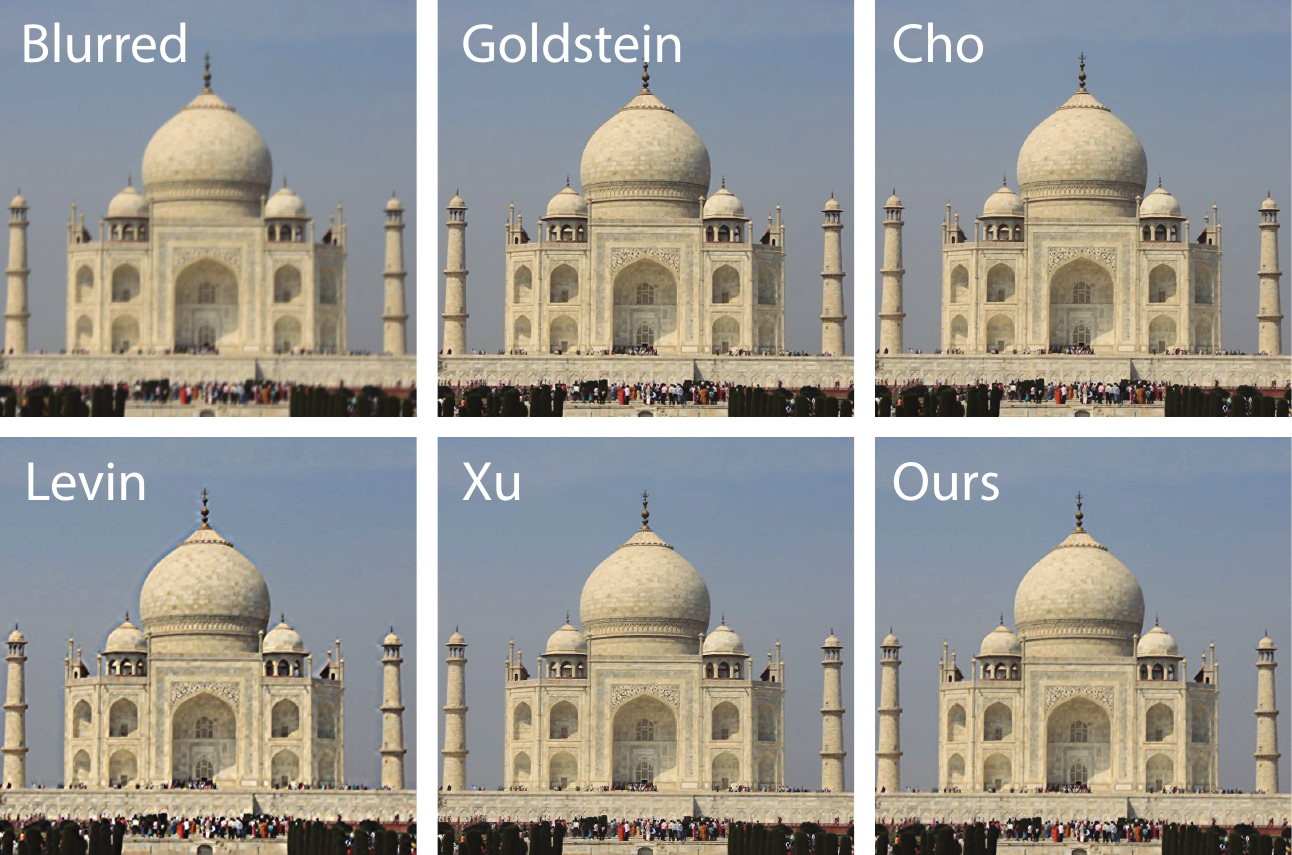}
\end{center}
\vspace{-4mm}
\caption{An example from Goldstein and Fattal \cite{Goldstein2012}.  We also include their result for comparison.}
\label{fig:real_goldstein}
\end{figure*}

In \ref{fig:real_goldstein}, we use an image from Goldstein and Fattal \cite{Goldstein2012}. The algorithm in that paper is based
on spectral arguments, and so does not fall under the variational or $MAP$ categories. Our method, Cho and Lee \cite{Cho09}
and Xu et al. \cite{xuunnatural} perform well. The output of Levin et al. \cite{levin2011efficient} results has artifacts around the edges.

\begin{figure*}[t]
\begin{center}
\includegraphics[width=3.5in]{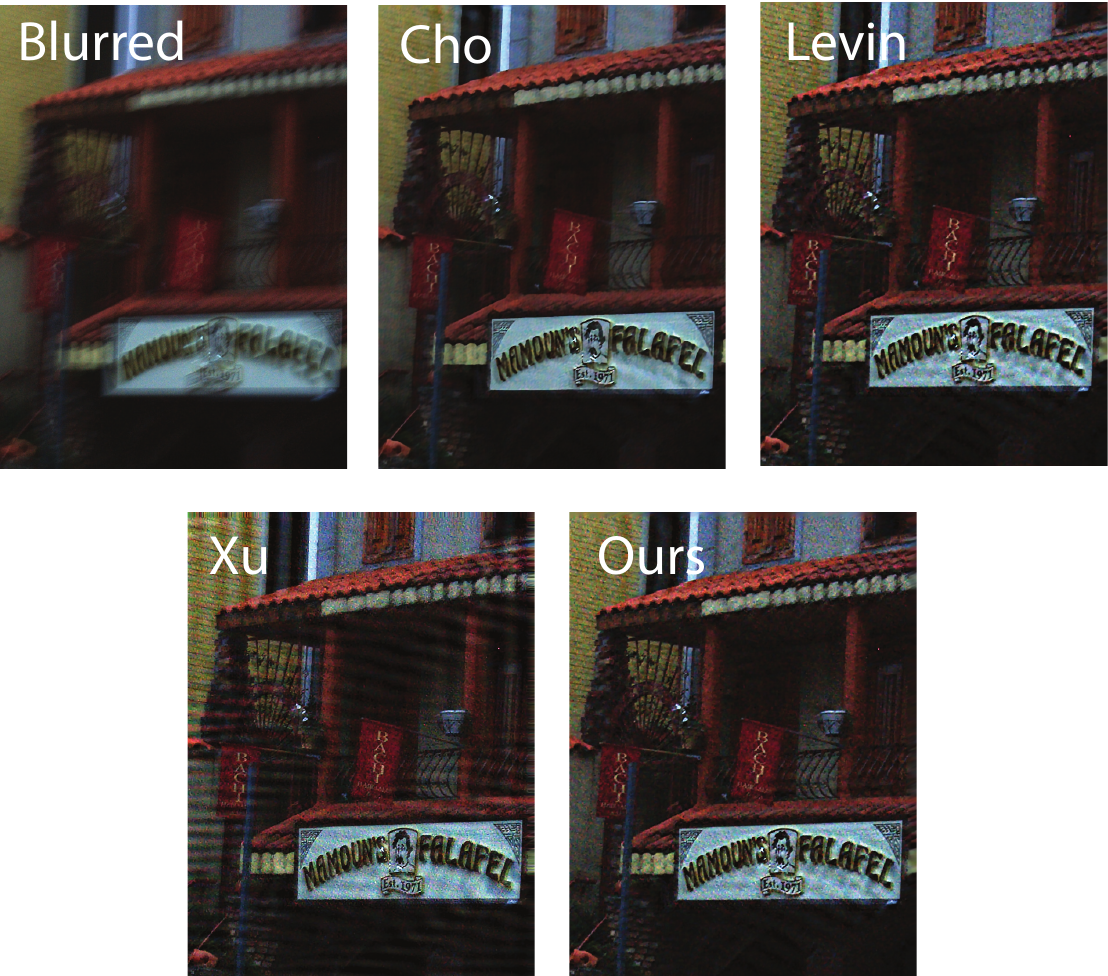}
\end{center}
\caption{A real-world example from Lin et al. \cite{chohandling} that exhibits blur and high noise levels. Note that unlike 
\cite{chohandling}, we find that the method of Levin et al. \cite{levin2011efficient} also performs well. The results of 
Xu et al. \cite{xuunnatural} exhibits significant ringing. }
\label{fig:real_zhong}
\end{figure*}

In a recent paper, Lin et al. \cite{chohandling} proposed a new algorithm to handle deblurring
in the case of very high noise levels. We show that our proposed algorithm is quite robust to such situations by
using an example from their paper (\ref{fig:real_zhong}). 
The algorithm of Xu et al. \cite{xuunnatural} produces significant
ringing. These could possibly be reduced by parameter adjustments, but no parameters are exposed in 
their executable. Note that unlike the conclusions of Lin et al. \cite{chohandling}, we find that 
the algorithm of Levin et al. \cite{levin2011efficient} works quite well on this example. 

The code of Wipf and Zhang \cite{wipf2013revisiting} is not available. However, we note that our method seems to perform as well as 
theirs on the dataset of Levin et al. \cite{Levin09}. Finally, by modifying the likelihood term using the ideas in Whyte 
et al. \cite{Whyte10}, 
our method can be extended to the case of blur due to camera in-plane rotation. Our code and test data is available
at \url{www.xxx.yyy}.

\section{Discussion}
%
%
%

In this paper, we have discussed a number of common properties of successful blind deconvolution
algorithms, with sparsity promotion being the most important. In spite of the good performance of 
existing methods, a number of open problems remain.

The original formulation \ref{eqn:bd} is non-convex, and alternating minimization schemes are only 
guaranteed to reach a local minimum. The use of a multi-scale pyramid improves the numerical 
convergence, but it is quite possible to get stuck in sub-optimal solutions even in that scenario.
These problems tend to be exacerbated in large images with many levels
in the pyramid, where errors from the
coarse to fine scheme may gradually accumulate. Therefore, other minimization 
strategies such as the convex programming based approach of Ahmed et al. \cite{ahmed2012blind} may prove
to be better initialization strategies than the multi-scale scheme.

Existing sparsity promoting schemes are not consistent estimators of the blurring kernel $\kor$
because as the size of the input $y$ increases, they are penalised by estimation errors on the $\xor$. 
Consistent estimators may be obtained by extracting stable geometric structures, 
using non-local regularisation terms, such as those presented in (\ref{eqn:our_wts}).
Highly oscillatory textures do not corrupt the estimation of $\kor$, thus showing that
sparsity can be highly efficient even 
when input images do not have sparse gradients.
Reweighting schemes provide efficient algorithms for that purpose, although 
their mathematical properties remain an open issue.

\appendix

\section{Proof of Proposition \ref{sparseprop}}
\label{proofsect}

Given $\Omega$, we define 
$\Omega_S = \{ i \,\,s.t.\,\, dist(i,\Omega) \leq S\}~,$
and we decompose the likelihood term as 
\begin{equation}
\label{tom1}
\| y - \tilde{x} \star k \|^2 = \restra{\| y - \tilde{x} \star k\|^2}{\Omega_S} + \restra{\| y - \tilde{x} \star k \|^2}{\Omega_S^c}~.
\end{equation}
Since $\restr{x}{\Omega_S^c}\equiv 0$, and $k$ has compact support
smaller than $S$, it results that 
\begin{equation*}
\| y - \tilde{x} \star k \|^2 = \restra{\| y - \tilde{x} \star k\|^2}{\Omega_S} + \restra{\| y  \|^2}{\Omega_S^c}~,
\end{equation*}
and hence 
{\small 
\begin{eqnarray}
\label{tom2}
\kest &=& \arg \min_k \| y - \tilde{x} \star k \|^2 + \lambda \| k \|^2 \nonumber \\
&=& \arg \min_k  \restra{\| y - \tilde{x} \star k\|^2}{\Omega_S} + \lambda \|k \|^2~.
\end{eqnarray}
}
Since $\sum_i \kest_i = \sum_i \kor_i = 1$ by construction,
we shall restrict ourselves to the subspace 
$\{ k \, ; \,  \langle k, {\bf 1} \rangle = 1\}$.
If $y=\xor \star \kor + n$ and $e = \xor - x$, it follows that
\begin{eqnarray*}
\kest &=& \arg \min_k  \restra{\| \xor \star (\kor - k) + n - e \star k \|^2}{\Omega_S} + \lambda \|k \|^2~. 
\end{eqnarray*}
By denoting by $A$ and $\tilde{A}$ the linear operators
$$A(y) = P_{\Omega_S} (\xor \star y)~,~\tilde{A}(y)=P_{\Omega_S}( e\star y)~,$$
it results from (\ref{tom2}) that 
{\small 
\begin{eqnarray*}
\kest &=& \left((A+ \tilde{A})^T(A + \tilde{A} ) + \lambda {\bf I}\right)^{-1}  [ (A + \tilde{A})^T A \kor + (A+\tilde{A})^T n ] \nonumber \\
&=& \left( \overline{A} + F \right)^{-1} \left( \overline{A} \kor + f \right)~,
\end{eqnarray*} }
with $\overline{A}=A^T A$,
$F=A^T \tilde{A} + \tilde{A}^T A + \tilde{A}^T \tilde{A} + \lambda {\bf I}$ and 
$f=\tilde{A}^T A \kor + (A+\tilde{A})^T n $.
Since $\delta>0$, it results that $\overline{A}=A^T A$ is invertible 
in the subspace of $0$-mean vectors.
Since
\begin{equation*}
\left( \overline{A} + F \right)^{-1} \left( \overline{A} \kor + f \right) = ({\bf 1} + \overline{A}^{-1} F)^{-1} \kor + \overline{A}^{-1} f~,
\end{equation*}
it follows that 
{\small
\begin{eqnarray*}
\| \kest - \kor \| &\leq& \| \left( {\bf I} + \overline{A}F \right)^{-1} - {\bf I} \| \|\kor \| + \delta \| f\|  \\
&\leq & \frac{\| \overline{A} F \|}{1 - \| \overline{A} F \| } \| \kor \| + \delta^{-1} (\epsilon \| \kor \| \gamma + (\gamma+\epsilon) \|n\|_\Omega) \\
&\leq & O(\max(\epsilon \delta^{-1/2} ,\epsilon \gamma \delta^{-1}, \lambda )) \|\kor\| + O( (\gamma+\epsilon) \delta^{-1} \|n\|_\Omega )~~\square~.
\end{eqnarray*}
}

\bibliographystyle{splncs}
\bibliography{unified_blind_deconv}
\end{document}